\documentclass{article}
\usepackage[english]{babel}
\usepackage[square,numbers]{natbib}

\usepackage{aaai21}

\usepackage[utf8]{inputenc} 
\usepackage[T1]{fontenc}    
\usepackage{hyperref}       
\usepackage{url}            
\usepackage{booktabs}       
\usepackage{amsfonts}       
\usepackage{nicefrac}       
\usepackage{microtype}      
\usepackage{xcolor}         
\usepackage{graphicx}
\usepackage{caption}
\let\SUP\textsuperscript

\title{Bias and Fairness on Multimodal Emotion Detection Algorithms}

\author {
    Matheus Schmitz\SUP{1},
    Rehan Ahmed\SUP{2}, 
    Jimi Cao\SUP{3}  \\
}
\affiliations {
    University of Southern California \\
    \{\SUP{1}mschmitz, \SUP{2}rehanahm, \SUP{3}jimicao\}@usc.edu
}

\begin{document}

\maketitle

\begin{abstract}
Numerous studies have shown that machine learning algorithms can latch onto protected attributes such as race and gender and generate predictions that systematically discriminate against one or more groups. To date the majority of bias and fairness research has been on unimodal models. In this work, we explore the biases that exist in emotion recognition systems in relationship to the modalities utilized, and study how multimodal approaches affect system bias and fairness.
We consider audio, text, and video modalities, as well as all possible multimodal combinations of those, and find that text alone has the least bias, and accounts for the majority of the models' performances, raising doubts about the worthiness of multimodal emotion recognition systems when bias and fairness are desired alongside model performance.
\end{abstract}

\section{Introduction}
Neural Networks are already ubiquitously used for classification and recognition in countless domains, including humans and their emotions. Yet in the rush to continuously advance those systems, researchers have not always paid attention to unequal development in the system capabilities \cite{mehrabi}. Oftentimes those systems are found to work better in certain demographic groups than in others, with the “others” category most often being minorities \cite{corbett}. As a result, theses systems can potentially show prejudice in their judgement and consequently exacerbate inequality.

Emotion Recognition is an important area of research to enable effective human-computer interaction. Human emotions can be detected using speech signal, facial expressions, body language, and electroencephalography (EEG) \cite{karray}. Our goal is to analyze and try to mitigate gender bias in current state-of-the-art transfer learning algorithms for different modalities including audio, text, and video. We define model bias as the difference in the predictive performance of a model across different groups. Throughout this work the groups considered are gender-based, that is, male and female.

\section{Literature Review}
A survey on bias and fairness in machine learning by Mehrabi et al. \cite{mehrabi} points to the various sub-areas of Machine Learning in which the bias thematic has been explored by researchers, and, among other findings, reveals that the majority of published work has been on unimodal models. In the realm of multimodal learning, work by Shen Yan et al. \cite{shenyan} focusing on personality assessments showed that different modalities display numerous bias patterns, and that the data fusion stage also introduces further model bias. Similarly, Ramakrishna et al. \cite{ramak} revealed how gender, race, and age can affect the way characters are portrayed in media. Buolamwini et al. \cite{gshades} found substantial disparities in the accuracy rate of classifying gender of dark-skin females in existing commercial gender classification systems.

Research by Yoon et al. \cite{yoon} worked on audio-text bimodal emotion recognition but there was no emphasis on assessing or mitigating bias. Tripathi et al. \cite{tripathi2018multi} expanded this line of research by exploring audio-text-video trimodal emotion recognition, but still leave out any considerations about bias and fairness. Meanwhile, Domnich and Anbarjafari \cite{domnich} worked on gender bias assessment in emotion recognition using unimodal learning from facial imagery. 

Given this paper analyzes bias, we need to have a definition of fairness and how to measure and compare multiple modalities in their fairness. Kleinberg et al. \cite{klien} have shown that it is simply not possible to satisfy all definitions of fairness at the same time. Hence here we opt to utilize the fairness definition proposed by Bellamy et al. \cite{bellamy}, which are composed of two metrics: statistical parity difference, and equality of opportunity difference \cite{fairnessmeasures}. 

Statistical parity is a fairness metric that measures if the prediction is independent of the protected attribute and can be calculated with the following equation: 
\begin{equation}
    Pr(\hat{y}|p) = Pr(\hat{y})
    \label{eq:statistical_parity}
\end{equation}

Equality of Opportunity is a fairness metric which measures if the prediction is conditionally independent to the protected attribute, given the value of predictor is true and can be calculated with the following equation:
\begin{equation}
    Pr(\hat{y}|y=1,p) = Pr(\hat{y}|y=1)
    \label{eq:equality_of_opportunity}
\end{equation}

As this literature review shows, individually there is prior work in multimodal emotion recognition, fairness in multimodal learning, and gender bias in machine learning. This work advances current research by exploring the intersection of these, namely: gender bias in state-of-the-art transfer learning models for emotion recognition, considering both unimodal as well as multimodal fusion architectures.

\section{Dataset}

We used the IEMOCAP dataset \cite{IEMOCAP}. This dataset contains both a scripted and an improvised section, and we opt to utilize both in our analysis so as to maximize our sample sizes during model training. In including acted data there is a risk the findings are not representative of day-to-day life, but given there are no large datasets of non-acted multimodal emotion recognition data covering audio, text, and video, we opt for IEMOCAP as a solution for covering all modalities that is often utilized for emotion recognition research (\cite{tripathi2018multi, kim2021emoberta, mirsamadi2017automatic, satt2017efficient, issa2020speech}). While it would be possible to utilize only the non-acted part of IEMOCAP, developing models on small data has numerous side-effects which would also result in questionable findings. For this reason we choose to utilize the full collection of samples available in IEMOCAP. The dataset originally contains 10039 utterances, with a class breakdown as per table \ref{tab:class_breakdown}. Those samples have are split roughly equally amongst 5 sessions, each containing one male and one female participant.

In IEMOCAP, "xxx" identifies samples for which there was no agreement amongst annotators, while "other" means annotators all agree on some emotion, but that emotion is not part of the predefined emotions for the dataset. On the basis of not being useful labels we remove all samples labeled as "xxx" and "other". On the basis of having too small of a sample size we remove the samples labeled as: "surprised", "fearful", "disgusted. Lastly, similar to previous work on the IEMOCAP dataset (\cite{Poria2016, Neumann2019, Sutherland2021}), we merge the "excited" and "happy" classes into a single class for which we use the "happy" label. As a result, the final dataset contains 7380 samples, with a class breakdown shown in table \ref{tab:class_breakdown}. All classes are balanced, with a roughly 50/50 gender split.

Utilizing the filtered data, we create the following train-validation-test split: Sessions 1-4 are training data, on Session 5 the even numbered utterances are validation data, and the odd numbered utterances are test data. This implies roughly a split of 80\% training, 10\% validation and 10\% test.

\begin{table}[h]
\centering
\small
\begin{tabular}{c c c} 
 \toprule
 \textbf{Emotion} & \textbf{Original Dataset} & \textbf{Filtered Dataset} \\
 \midrule
 xxx          & 24.9 \% &  - \\ 
 frustrated   & 18.4 \% &  25.1 \% \\ 
 neutral      & 17.0 \% &  23.1 \% \\ 
 angry        & 11.0 \% &  14.9 \% \\ 
 sad          & 10.8 \% &  14.7 \% \\ 
 excited      & 10.3 \% &  - \\ 
 happy        &  5.9 \% &  22.2 \% \\ 
 surprised    &  1.1 \% &  - \\ 
 fearful      &  0.4 \% &  - \\ 
 other        &  0.3 \% &  - \\ 
 disgusted    &  0.2 \% &  - \\ 
 \midrule
 Total samples  & 10039 & 7380 \\
 \bottomrule
\end{tabular}
\caption{Class breakdown on the dataset used}
\label{tab:class_breakdown}
\end{table}

\section{Methods}

\subsection{Feature Extractions}
We chose three different state of the art deep learning models for each of the three modalities available in the dataset: audio, video, and text. We leverage each models to extract embeddings from the data. 

\begin{figure}[h]
    \centering
    \includegraphics[width=0.95\linewidth]{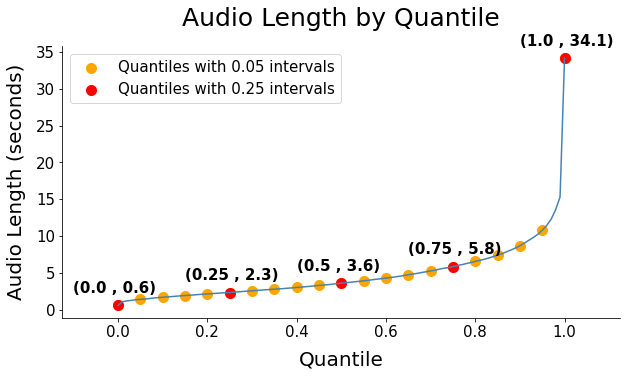}
    \caption{Utterance length by quantile}
    \label{fig:Audio_Utterance_Quantiles}
\end{figure}

\textbf{Audio.}
We leverage the utterance level data that comes in IEMOCAP and generate sliding time windows from which embeddings can be extracted. Utterance lengths vary from 0.6 seconds up to 34.1 seconds, with the 80th percentile being 6 seconds, and hence that is the chosen length to which all audios were standardized. The full utterance length distribution is shown in figure \ref{fig:Audio_Utterance_Quantiles}. For larger files the middle section of the data is kept, and for smaller files we employ front-padding, which we found to greatly improve model performance in comparison to back-padding. The audio is downsampled to 16kHz, and we utilize a 25 millisecond sliding window with a 10 millisecond step size, thus the resulting embeddings are of shape (timesteps, features) = (600, 400).

The unimodal model is utilized to compare five different feature extraction approaches: MFCC, Mel-Spectrogram, Wav2Vec2 \cite{Wav2Vec2}, TRILL \cite{TRILL} and WavLM\cite{wavlm}. Unimodal test dataset F1-Scores are as follows: MFCC (24.9\%), Mel-Spectrogram (26.3\%), Wav2Vec (28.1\%), TRILL (27.8\%), WavLM (28.6\%). Based on this analysis WavLM is the chosen embedding model. 


\textbf{Text.}
We use EmoBERTa, a technique proposed by Kim et al \cite{kim2021emoberta} that is a simple yet expressive scheme of emotion recognition in conversation task. It starts with RoBERTa \cite{RoBERTaA} for sequence classification and adds a randomly initialized linear layer with the softmax nonlinearity to the last layer of the pretrained model.

The authors propose multiple models, such as including past and future utterances for speaker state aware emotion recognition. Our work uses the pre-trained EmoBERTa-base model without any past nor future utterances, to keep it consistent with audio and video modality techniques in using only the utterance itself to recognize emotion.

The model was then further fine-tuned on our filtered IEMOCAP train split for 30 epochs using the training configurations proposed in the paper with 12 attention heads, 12 hidden layers, a Vocab Size of 50265 and 512 Max Position Embeddings. To use for fusion in multimodal techniques, we then extracted sentence-level embeddings by using the approach proposed by McCormick and Ryan (2019) \cite{bertwordembeddings}. For each individual token in the sentence, we summed the vectors from the last four layers. Next, we averaged all the extracted token vectors in the sentence to get a standardized sentence-level embedding for each sentence.

\textbf{Video.}
FaceNet \cite{facenet} is the approach chosen to extract embeddings from the video files. We split the videos based on the time frame of the utterances in the audio section. This resulted in video spanning from 18 to 1024 frames with the 90th percentile being 263 frames, at 30 frames per second. IEMOCAP's videos contain two speakers at any given moment. The video files contain labels regarding in which halves of the video the male and female subjects were, and also on which gender was speaking at any given utterance. Using these information we were able to identify and isolate the active speaker. Then, using MTCNN \cite{MTCNN}, a pre-trained neural network for face detection, we detected faces and extracted 512-features-long embeddings for the detected face through FaceNet for each frame of a video. For frames where faces were not detected, an empty array was used in its place instead. Finally, for videos with less than 263 frames, we pre-padded the frames. For videos with more than 263 frames, we kept the middle section.

\subsection{Model Training}
\begin{figure*}[h]
    \centering
    \includegraphics[width=0.8\linewidth]{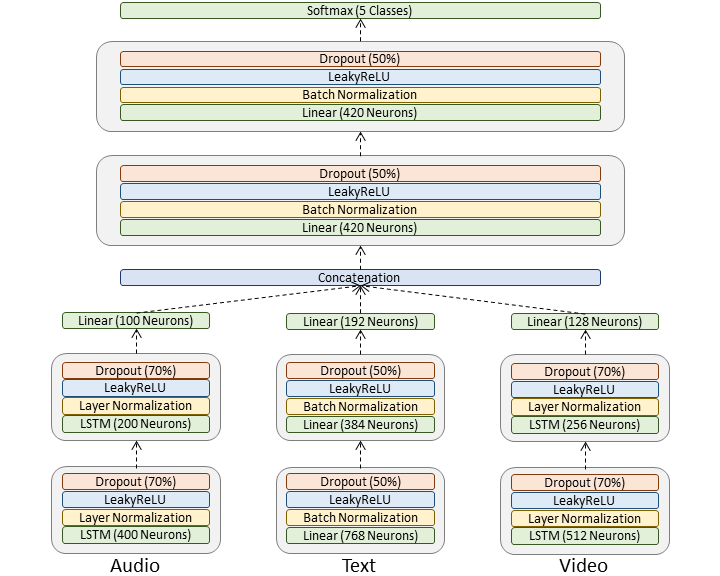}
    \caption{Architecture for the neural network utilized}
    \label{fig:Network_Architecture}
\end{figure*}

We chose to utilize a standardized model architecture to make comparisons more direct. The utilized neural network structure is shown in figure \ref{fig:Network_Architecture}. With this structure, we train one model with each possible combination of modalities. For unimodal and bimodal training, the modalities not being utilized simply forward a vector of zeros to the fusion layer. The chosen optimizer is AdaBelief \cite{AdaBelief} with a learning rate of 1e-3, and the loss function is Categorical Cross Entropy. All layers have a LeakyReLU activation, except for the final layer which is a Softmax.

We employ early stopping to terminate training for any model that doesn't see a reduction in validation set loss for 20 sequential epochs. All models are allowed to train until the early stopping trigger halts training, and the models are then restored to the state (weights) they had on the best epoch, as judged by validation loss.

The specific choices for normalization and dropout on the individual modalities' branches were derived experimentally on the unimodal models associated with each modality, during this process a Softmax layer was utilized in place if the pre-concatenation linear layer. The scores from those reduced models were not utilized for evaluation, merely for architecture design. Once the individual modalities' branches were locked, the post-concatenation section of the architecture was derived in a similar manner considering the trimodal model. When the final architecture was locked in all modality variations were trained on the same neural network structure.

\section{Results}

\subsection{Model Performance}
We compared models across three different metrics: Accuracy, F1-Score, and ROC-AUC. Table \ref{tab:model_performance} summarizes the results. We can observe that across all metrics the "audio + text" model was the highest performing one. The unimodal "video" model had the worst performance across all metrics, which might explain why adding it to the best model results in a worse trimodal model.

\begin{table}[h]
\centering
\small
\begin{tabular}{c c c c} 
 \toprule
 \textbf{Model} & \textbf{Accuracy} & \textbf{F1} & \textbf{ROC}\\
 \midrule
 audio                   & 0.2864  &  0.2704  &  0.5780  \\ 
 text                    & 0.6531  &  0.6518  &  0.7901  \\ 
 video                   & 0.2296  &  0.1314  &  0.5087  \\ 
 \textbf{audio + text}   & \textbf{0.6580}  &  \textbf{0.6564}  &  \textbf{0.7940}  \\ 
 audio + video           & 0.3012  &  0.2676  &  0.5895  \\ 
 text + video            & 0.6531  &  0.6511  &  0.7883  \\ 
 audio + text + video    & 0.6519  &  0.6539  &  0.7898  \\ 
\bottomrule
\end{tabular}
\caption{Modality performance on the test dataset}
\label{tab:model_performance}
\end{table}

\subsection{Bias and Fairness}
We used a number of different metrics to compare how model performance varied across the two gender groups.

Table \ref{tab:Fairness_Measures_Comparison1} shows how each modality performs on: (1) Average F1 Score difference, i.e. the difference across genders in F1 Score; (2) Average Equality of Opportunity (EO) difference, i.e. the average difference across genders in how conditionality independent the prediction is to the protected attribute; and (3) Average Statistical Parity (SP) difference, i.e. the average difference across genders in selection rates of each emotion. For all these metrics lower is better. We can see that unimodal audio and text are the least biased across all measures. Among unimodal approaches Video is significantly more biased. Considering bimodal models, fusing audio and video significantly increases bias in the system by increasing statistical parity and equality of opportunity.

\begin{table}[h]
\centering
\small
\begin{tabular}{c c c c} 
 \toprule
 \textbf{Model} & \textbf{F1} & \textbf{EO} & \textbf{SP}\\
 \midrule
 audio                   & 1.36  &  1.82  &  \textbf{1}  \\ 
 text                    & \textbf{1.01}  &  \textbf{1.43}  &  2  \\ 
 video                   & 4.63  &  2.32  &  3  \\ 
 audio + text            & 1.86  &  2.21  &  2  \\ 
 audio + video           & 1.91  &  2.22  &  2  \\ 
 text + video            & 2.77  &  3.24  &  6  \\ 
 audio + text + video    & 2.14  &  2.68  &  2  \\ 
\bottomrule
\end{tabular}
\caption{Comparison of fairness measures across modalities. F1 = Average F1 Score difference between genders. EO = Average Equality of Opportunity difference between genders. SP = Average Statistical Parity difference between genders.}
\label{tab:Fairness_Measures_Comparison1}
\end{table}

\begin{figure*}[h]
    \centering
    \includegraphics[width=0.9\linewidth]{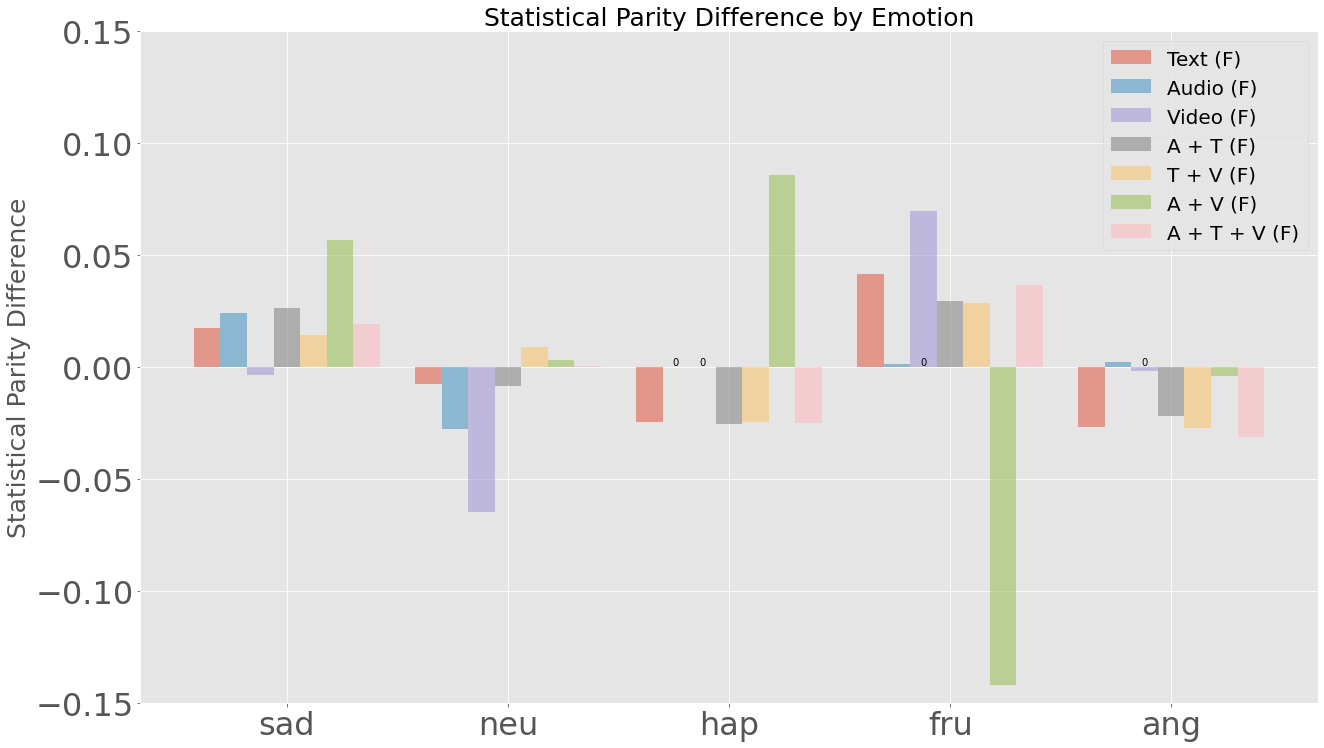}
    \caption{Statistical Parity difference by emotion and modality}
    \label{fig:Fairness_Measures_Comparison2}
\end{figure*}

In figure \ref{fig:Fairness_Measures_Comparison2}, we compare Statistical Parity across each emotion. The closer to 0, the fairer the model is towards both groups. Being close to -1 means the selection rate for females is higher while +1 means selection rate for males is higher. Using this measure for fairness, we could see Text and audio were the least biased unimodal models while video was the most biased according to both the fairness measures, especially for happiness and frustration. The trimodal approach yielded decent results, similar to text and audio unimodal approaches.

Results for Equality of Opportunity are highly similar, hence we opted to place them in the appendix section.

\section{Conclusion}
We surveyed different unimodal and multimodal approaches for emotion recognition, and showed that using text alone was superior to other approaches at emotion recognition by achieving a good balance between accuracy and fairness measures. Audio, although significantly worse than text, was still a lot better than video, as the latter wasn't able to obtain good accuracy while also having the worst performance in fairness measures. Amongst bimodal models, an audio + video fusion is the best performing model, and the addition of video results in a worse trimodal model in both predictive performance and fairness. We note that compared to text alone, the F1 Score gain from the audio-video bimodal model is rather small in comparison to the more accentuated worsening of fairness metrics.

Additionally, this work shines light on how different modalities varied on recognizing different emotions. Both unimodal and multimodal models which included text features were worse at detecting anger in females compared to males, displaying a gap in F1-Score of roughly 10\%, as shown in the appendix. When fusing all the modalities, we observed that the trimodal model's performance was similar to text, failing to improve significantly on either accuracy or fairness.

The results presented give raise to one important ethical dilemma: One the one hand fusing modalities can potentially improve model performance, on the other hand adding modalities makes people more identifiable. This is relevant given the observation that text alone drives nearly all the model performance, while being the least biased. The latter arguably is a by-product of the fact that among text, audio, and video, text contains the least cues about someone's gender. This combination creates the predicament of deciding whether the addition of audio and video data to emotion recognition models is warranted in face of the small performance boost and the disproportional worsening of gender bias.

\subsection{Future Work}
Without the aid of more complex training techniques, in this paper we see the performance of models that do not include text is low (the random guess performance for 5 classes would be 20\%). Hence there remains the possibility that part of the bias currently observed for non-text modalities, especially video, is a result of their poor learning ability overall. For this reason we recommend that future work emphasizes raising performance for audio and video, by, for example, incorporating dataset fusion, data augmentation, and other techniques that reduce over-fitting and increase the model's learning capacity, as better performing models would enable a clearer isolation of factors between simple overall poor performance and bias.

We also observe a peculiarly consistent effect where all model variations that include text data show a significant performance gap in predicting anger for men and women. The consistency of this tendency through all fusion models makes it unlikely to be a one time fluke. Yet we cannot, based on the current work, identify whether this is associated with the dataset used, with differences in vocabulary choices between genders, with the performance gap between modalities, or with other reasons. Thus such quandary provides an interesting avenue for further exploration.



\bibliography{biblio}

\clearpage
\LARGE{Apendix}

\begin{center}
    \centering
    \includegraphics[width=2\linewidth]{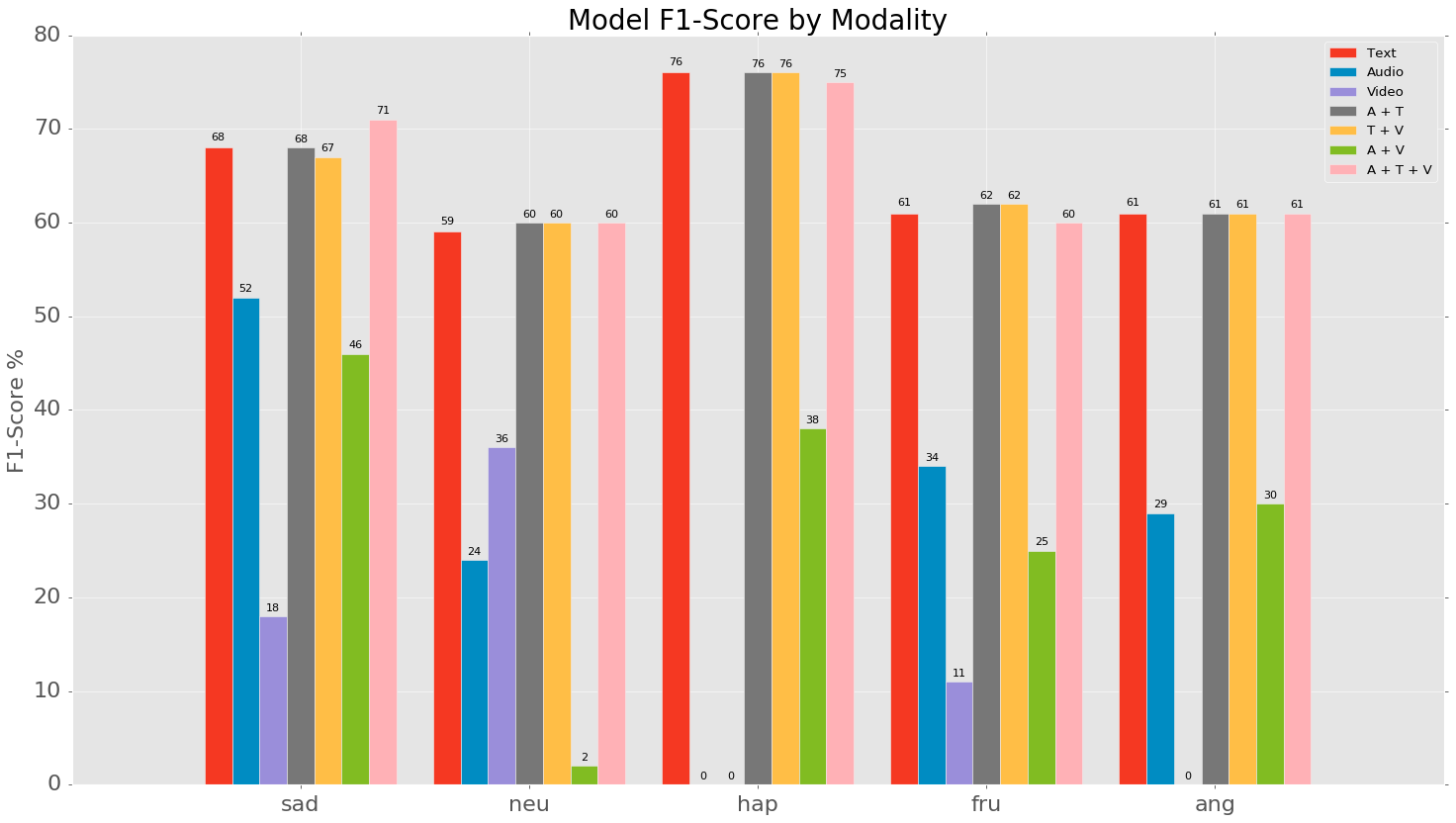}
    \label{fig:F1_Emotion_Modality}
\end{center}

\begin{center}
    \centering
    \captionsetup{justification=centering}
    \includegraphics[width=2\linewidth]{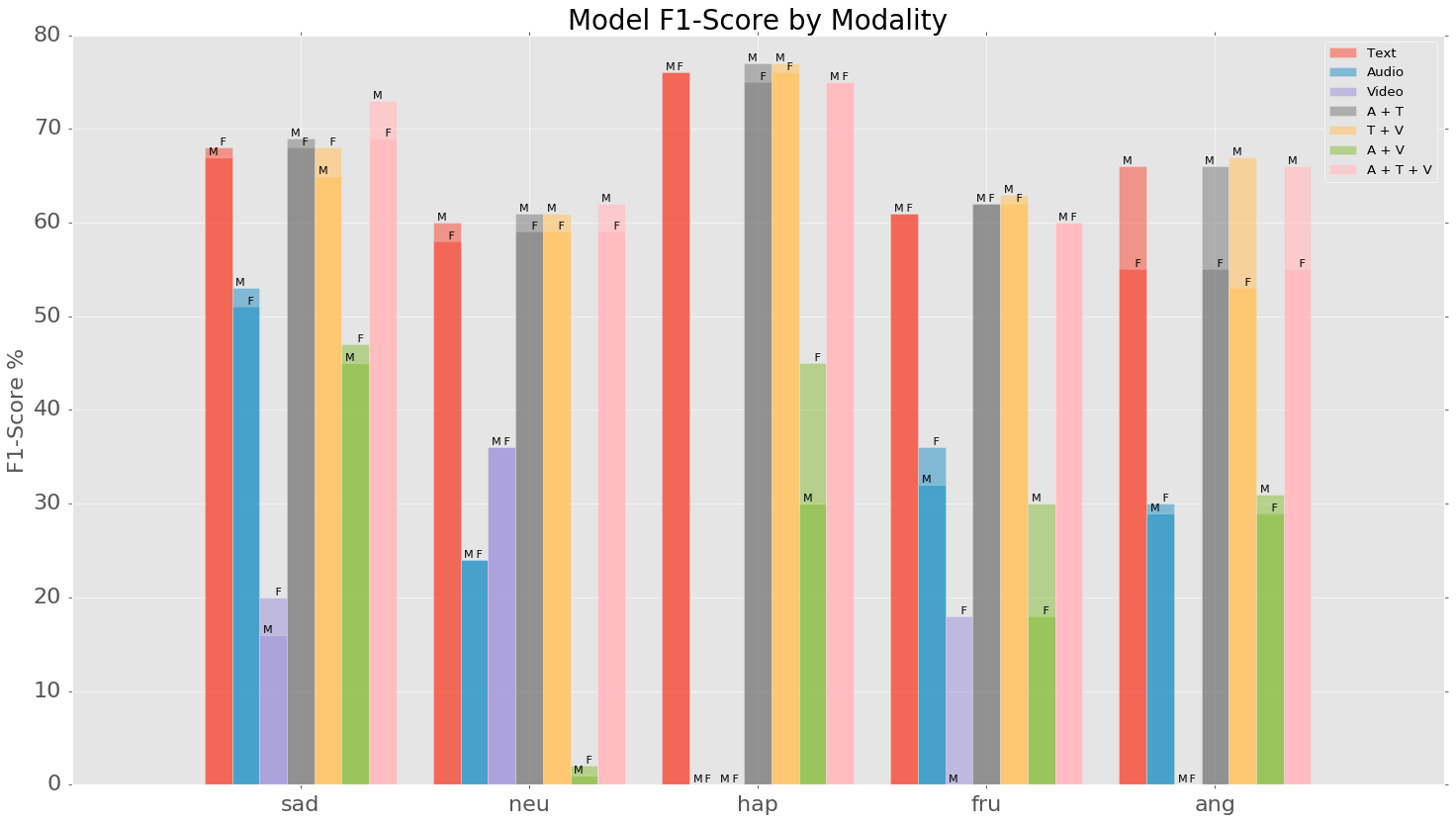}
    \label{fig:F1_Emotion_Modality_Gender}
\end{center}

\clearpage
\begin{center}
    \centering
    \captionsetup{justification=centering}
    \includegraphics[width=2\linewidth]{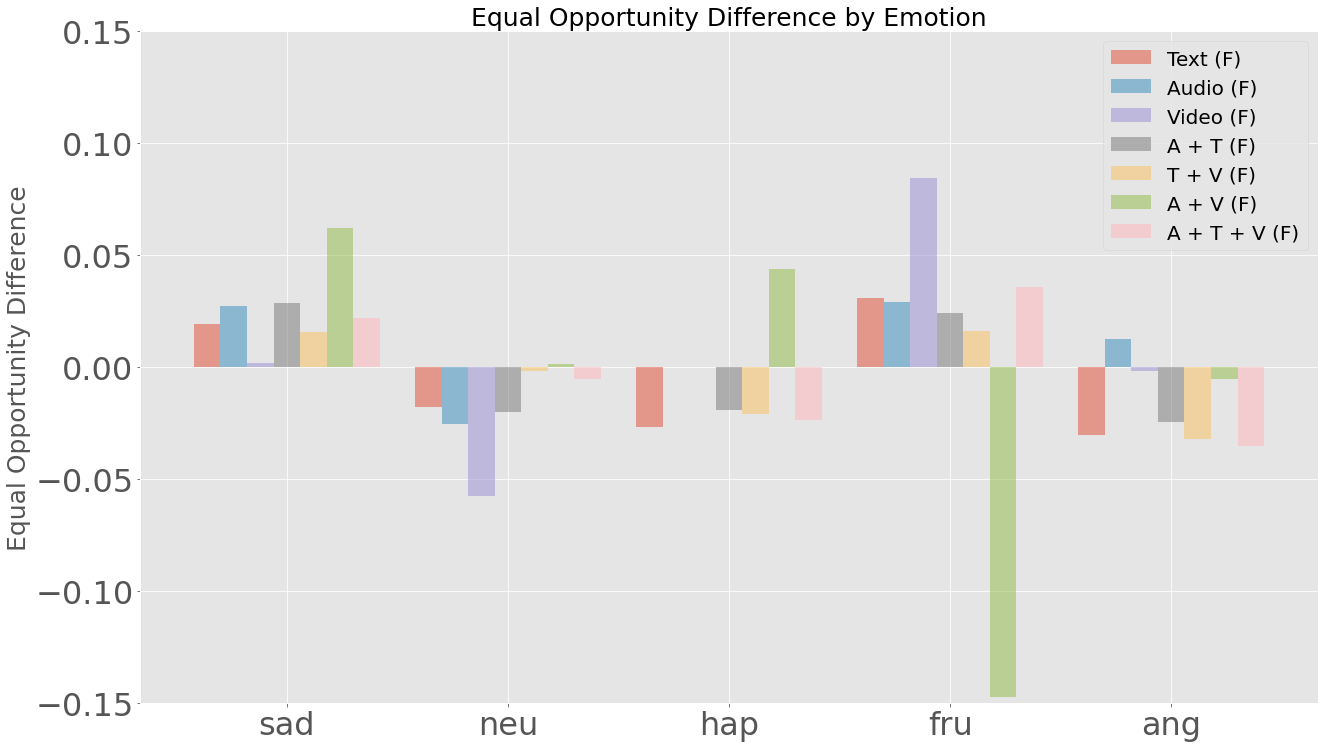}
    \label{fig:eq_graph}
\end{center}

\end{document}